\pdfoutput=1
\documentclass[11pt]{article}

\usepackage[]{acl}
\usepackage[normalem]{ulem}
\usepackage{colortbl}
\usepackage[ruled,vlined,linesnumbered]{algorithm2e}
\usepackage{geometry}
\usepackage{times}
\usepackage{latexsym}
\usepackage{amsfonts}
\usepackage{amsmath}
\usepackage{pgfplots}
\usepackage{subfigure}
\usepackage{multirow}
\usepackage{xcolor}
\usepackage{todonotes}
\definecolor{color1}{RGB}{9,147,150}
\definecolor{color4}{RGB}{0,18,25}
\definecolor{color3}{RGB}{238,155,0}
\definecolor{color2}{RGB}{174,32,18}

\usepackage[T1]{fontenc}
\usepackage[utf8]{inputenc}

\usepackage{microtype}

\newcommand{\methodname}{WSPAlign}
\newcommand{\methodnamesix}{{\methodname}-M6}
\newcommand{\methodnameone}{{\methodname}-E}

\title{\methodname: Word Alignment Pre-training via Large-Scale\\ Weakly Supervised Span Prediction}

\author{Qiyu Wu$^{1}$,
        Masaaki Nagata$^{2}$,
        Yoshimasa Tsuruoka$^{1}$\\
$^{1}$\normalfont{The University of Tokyo}, Tokyo, Japan \\
$^{2}$\normalfont{NTT Communication Science Laboratories, NTT Corporation, Kyoto, Japan} \\
$^{1}$\normalfont{\texttt{\{qiyuw, yoshimasa-tsuruoka\}@g.ecc.u-tokyo.ac.jp}} \\
$^{2}$\normalfont{\texttt{masaaki.nagata@ntt.com}}
}

\begin{document}
\maketitle

\begin{abstract}
Most existing word alignment methods rely on manual alignment datasets or parallel corpora, which limits their usefulness.
% \nagata{I think you'd better not mention "low resource" many times because there are no experiments on low resource in this paper.}
Here, to mitigate the dependence on manual data,
% we propose a novel weakly supervised pre-training approach \methodname, without requiring any manual dataset or parallel corpus.
% Considering the trade-off between quality and quantity,
we broaden the source of supervision by relaxing the requirement for correct, fully-aligned, and parallel sentences. Specifically, we make noisy, partially aligned, and non-parallel paragraphs.
We then use such a large-scale weakly-supervised dataset for word alignment pre-training via span prediction.
% The approach is named \methodname.
Extensive experiments with various settings empirically demonstrate that our approach, which is named \methodname, is an effective and scalable way to pre-train word aligners without manual data.
When fine-tuned on standard benchmarks, \methodname~ has set a new state of the art by improving upon the best supervised baseline by \textbf{3.3\textasciitilde6.1} points in F1 and \textbf{1.5\textasciitilde6.1} points in AER
% \nagata{you need units like "3.3-6.1 points"}
% \nagata{It is better to state improvement in absolute percentage points than relative percentages.}
. Furthermore, \methodname~also achieves competitive performance compared with the corresponding baselines in few-shot, zero-shot and cross-lingual tests, which demonstrates that \methodname~is potentially more practical for low-resource languages than existing methods.
\footnote{The source codes are publicly available at \href{https://github.com/qiyuw/wspalign}{https://github.com/qiyuw/wspalign} (for pre-training) and \href{https://github.com/qiyuw/wspalign.infereval}{https://github.com/qiyuw/wspalign.infereval} (for inference and evaluation).}
\end{abstract}

\section{Introduction}
\label{sec:intro}
\begin{figure*}[ht!]
    \centering
    \includegraphics[width=\textwidth]{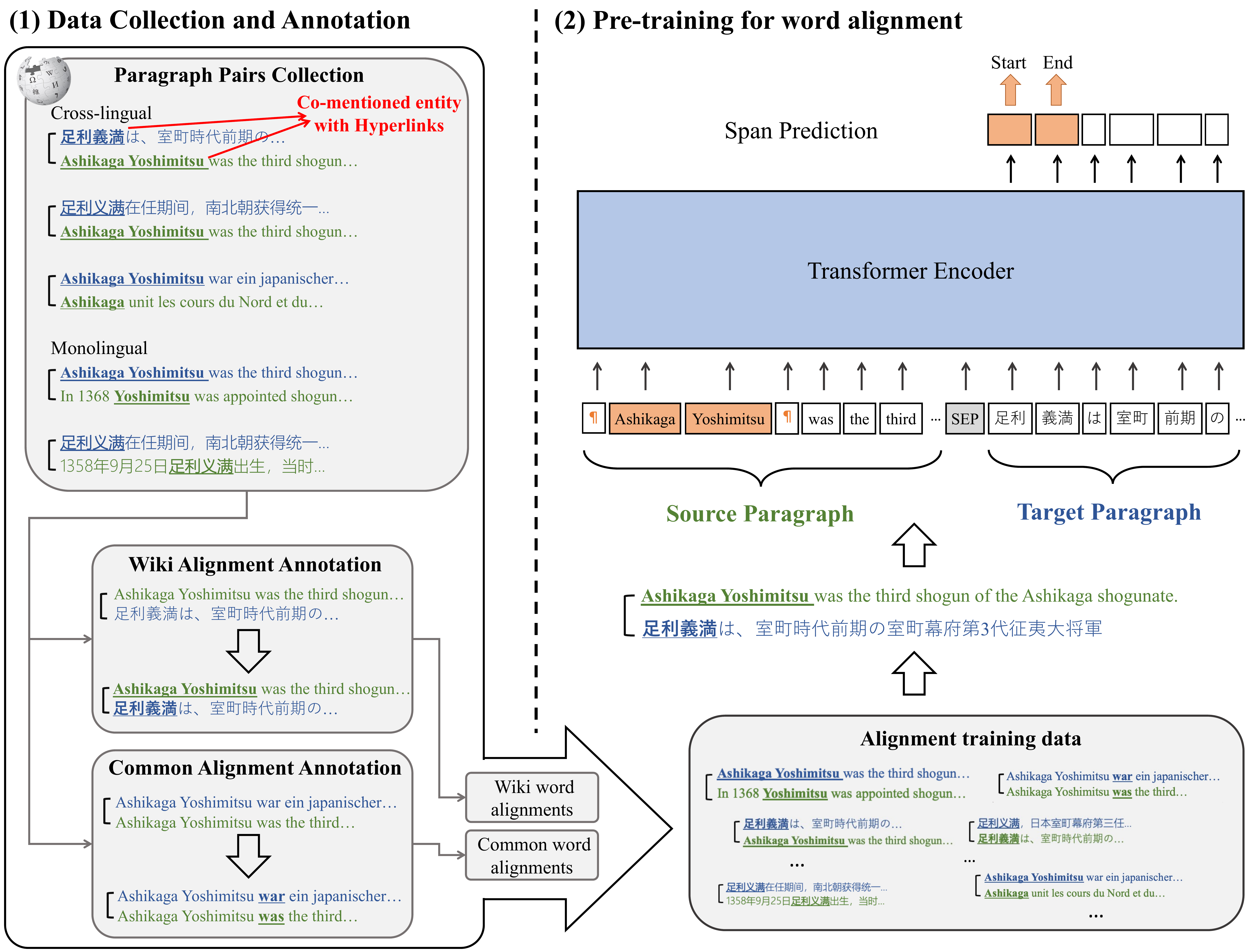}
    \caption{Framework of \methodname. Paragraphs are all collected from Wikipedia. We first collect paragraph pairs in which two paragraphs contain an identical language-agnostic entity.
Note that the paragraph pairs can be cross-lingual or monolingual, depending on the downstream application goals. 
Then, we automatically annotate word alignments for common words and wiki words separately and combine them together to make the final dataset. Lastly, the model is pre-trained on the above collected weakly-supervised datasets via the span prediction task.}
% \vspace{-2mm}
    \label{fig:s3a}
\end{figure*}

Word alignment, which aims to align the corresponding words in parallel texts, is a fundamental Natural Language Processing (NLP) task that was originally developed as an important supporting task for statistical machine translation. While deep end-to-end models have become the mainstream solution for machine translation, word alignment is still of great importance in many NLP scenarios, such as projecting linguistic annotations~\cite{david2001inducing} and XML markups~\cite{hashimoto2019high}, post-editing for detecting problem of under-translation~\cite{tu2016modeling}, and enforcing pre-specified terminology constraints in translation~\cite{song2019code}. Besides, word alignment can also improve the cross-lingual language pre-training~\citep{chi2021improving}.

% \qiyu{(Double-check needed) Previous word aligners are typically learned based on the performance on the performance of machine translation\cite{}. Recent efforts significantly improve previous word alignment tools with attempts to explore the supervised methods with pre-training on parallel sentences~\cite{garg2019jointly,stengel2019discriminative}, or unsupervisedly to exploit the information within the multi-lingual Pre-trained Language Models(PLMs)~\cite{JaliliSabet2020SimAlignHQ,dou2021word}.}
% Particularly, \citet{nagata2020supervised} proposed to formalize this problem as supervised span prediction using PLMs like BERT~\cite{devlin2019bert}, which had set a new state-of-the-art on multiple standard benchmarks without the need for parallel corpus.

However, most existing word alignment methods rely on either manual alignment datasets or parallel corpora for training, which weakens their usefulness because of the limiting accessibility of data.
% Detailed related works are introduced in \S\ref{sec:related}.
An additional weakness with requiring manual data is the generalization ability because deep models trained on a dataset can fail on other datasets. Therefore, these existing approaches are also limited in terms of potential cross-lingual use. 
% Therefore, we should mitigate the dependence of word aligners on manual datasets, especially for low-resource languages.
% Recently in the field of NLP, large Pre-trained Language Models (PLMs), e.g., GPT-3~\cite{brown2020language}, have shown remarkably zero-shot performance on various tasks without requiring any task-specific datasets.
% \nagata{The first two sentences in this paragraph could be moved to related works if you focuss on weakly supervion.}
On the other hand, recent studies~\cite{mahajan2018exploring,kolesnikov2020big,chen2021gigaspeech,galvez2021people,radford2022robust} in various fields leverage weak supervision signals in large-scale data available on the web for pre-training, which is a promising alternative to training on manual data. 

% Although understudied so far for word alignment, recent works in other filed such as computer vision~\cite{mahajan2018exploring,kolesnikov2020big} and speech recognition~\cite{chen2021gigaspeech,galvez2021people,radford2022robust} have shown that larger but weakly supervised datasets surpass manually collected gold-standard datasets in terms of robustness and generalization of models.

Inspired by this, we propose to utilize Wikipedia and multi-lingual Pre-trained Language Models (PLMs) to make large-scale word alignment supervision for pre-training via span prediction.
We broaden the source of supervision by relaxing the requirements for \textbf{correct} (manually made), \textbf{fully-aligned} (all words in a sentence pair are annotated), and \textbf{parallel} sentences.
Specifically, we make \textbf{noisy} (automatically made), \textbf{partially-aligned}, and \textbf{non-parallel} paragraphs (or monolingual paragraph pairs).
% \nagata{You should precisely state how you relaxed the requirements in Introduction as well as Abstract.}
We make automatic partial alignment between non-parallel sentences from either co-mentions
\footnote{Co-mention means two paragraphs mention an identical entity.}
of entities obtained through Wikipedia hyperlinks or alignments of common words based on the similarity of contextual word embeddings.

We name our method \methodname, which is short for \textbf{\uline{W}}eakly \textbf{\uline{S}}upervised span \textbf{\uline{P}}rediction pre-training for word \textbf{\uline{Align}}ment.
With weak supervision, we are potentially able to scale the pre-training data up to millions of paragraph pairs in hundreds of languages.
For instance, we made tens of millions of paragraph pairs and sampled a dataset with 2 million pairs in experiments, far more than 5,000 training examples in the existing benchmark dataset.
With no requirement for manual datasets, our pre-training approach makes word aligners more practical. Extensive experiments provide empirical evidence for \methodname's effectiveness in zero-shot, few-shot and supervised settings. We also conduct monolingual pre-training to test \methodname's cross-lingual ability.

% \nagata{Weakly Supervised span Prediction pre-training for word Alignment?}

% Through extensive experiments on both multi-lingual and monolingual pre-training with various downstream settings, we empirically demonstrate that \methodname~is an effective and scalable way to learn word aligners without the requirement for any manual parallel corpus and gold labels, especially for low-resource languages because of the competitive performance in the few-shot and zero-shot settings.
% For instance, when fine-tuning on standard benchmarks, \methodname~ has set the new state-of-the-art by improving the supervised baseline averagely by \textbf{by 3.3\textasciitilde 6.1} on F1 and \textbf{1.5\textasciitilde 6.1} on AER. In the few-shot setting, \methodname~ achieves competitive performance compared with the supervised baseline. Besides, \methodname~ also shows practical zero-shot ability by outperforming the unsupervised baseline that does not use parallel corpus.\nagata{This paragraph could be removed because it says same thing as Abstract.}
\section{Related Work}
\label{sec:related}
\paragraph{Word Alignment}
% \qiyu{(Double-check needed) Previous word aligners are typically learned based on the performance on the performance of machine translation\cite{}. Recent efforts significantly improve previous word alignment tools with attempts to explore the supervised methods with pre-training on parallel sentences~\cite{garg2019jointly,stengel2019discriminative}, or unsupervisedly to exploit the information within the multi-lingual Pre-trained Language Models(PLMs)~\cite{JaliliSabet2020SimAlignHQ,dou2021word}.}
Recent word aligners based on pre-trained language models,  such as SimAlign~\cite{JaliliSabet2020SimAlignHQ}, AWESoME~\cite{dou2021word} and SpanAlign~\cite{nagata2020supervised,chousa-etal-2020-spanalign}, have significantly outperformed previous word aligners based on statistical machine translation, such as Giza++~\cite{och2003systematic} and FastAlign~\cite{dyer2013simple}. 
SimAlign is an unsupervised word aligner based on the similarity of contextualized word embeddings.
AWESoME and SpanAlign are supervised word aligners that are trained on parallel corpora and manual word alignments, respectively.
Particularly, \citet{nagata2020supervised} proposed to formalize this problem as supervised span prediction using PLMs like BERT~\cite{devlin2019bert}, which had set the new state of the art on multiple standard benchmarks without the need for parallel corpora. Inspired by this, we take span prediction as our pre-training objective in this paper.

% However, most above representative works rely on either manual alignment datasets or parallel corpus, which weakens their usefulness in low-resource languages because of the limiting accessibility of data.
% In this work, we propose pre-training word aligners without any manual datasets and parallel sentences to mitigate this problem.\nagata{This paragraph could be removed because it says the same thing in Introduction. Why not move the remaining parts before Conclusion?}
% \vspace{-1mm}
\paragraph{Weakly Supervised Pre-training}
% In this paper, we try to mitigate the dependence of word aligners on manual datasets.
Recent PLMs in the field of NLP, e.g., GPT-3~\cite{brown2020language}, have shown remarkable zero-shot performance on various tasks without requiring any task-specific datasets.
Although understudied so far for word alignment, recent studies in other fields such as computer vision~\cite{mahajan2018exploring,kolesnikov2020big} and speech recognition~\cite{chen2021gigaspeech,galvez2021people,radford2022robust} have shown that weakly supervised but larger datasets surpass manual ones with gold labels in terms of robustness and generalization of models. This suggests that large-scale weakly-supervised pre-training is a promising alternative to manually collected supervised datasets or parallel corpora.
% \nagata{I think this paragraph can be merged into Introduction.}

\section{Appraoch}
\label{sec:method}
% The texts are all collected from Wikipedia paragraphs. We first collect paragraph pairs (not necessarily parallel) in which two paragraphs contain an identical language-agnostic entity.
% Note that the paragraph pairs can be cross-lingual or monolingual, depending on the downstream application goals. 
% % Then, we can directly make alignments for entity words using the hyperlink span in the paragraph.
% % For the common words, we take the same idea in~\cite{JaliliSabet2020SimAlignHQ}, in which we make the alignments based on the contextual word embeddings generated by PLMs.
% Then, we automatically annotate word alignments for \textit{common words} and \textit{wiki words} separately and combine them together to make the pre-training dataset. 
% Lastly, the model is pre-trained on the above collected weakly-supervised datasets via the span prediction~\cite{nagata2020supervised} task.
% The detailed process of data collection and pre-training method will be introduced in the following subsections.
% which begins with the preliminary word alignment via span prediction, followed by the proposed \methodname, including automatic data collection, alignments annotation, and the pre-training task.
% \nagata{The description of outline can be removed.}

\subsection{Background}
\label{sec:spanalign}
% Word alignment aims to align corresponding words in parallel texts.\nagata{You can remove this sentence and start with "We investigate the possibility of word alignment based on span prediction because it achieves the state-of-the-art if there is a manuall word alignment data."}
We investigate the possibility of word alignment based on span prediction because it is state-of-the-art when manual alignment data is available.
\citet{nagata2020supervised} proposed to frame word alignment as a SQuAD-style span prediction task~\cite{rajpurkar2016squad}.
In SQuAD-style question answering, given a \textit{context} extracted from a Wikipedia paragraph and a \textit{question}, the goal is to predict the answer span within the context based on the given question.
Word alignment can be framed similarly, as shown in the top-right part in Figure~\ref{fig:s3a}.
Given a source sentence with a source token specified by the special token ¶, the goal is to predict the aligned tokens in the target sentence.

Formally, given a source sentence $X=x_1,x_2,...,x_{|X|}$ consisting of $|X|$ characters, a source token $X_{ij}=x_i...x_j$ that spans $(i,j)$ in the source sentence, a target sentence $Y=y_1,y_2,...,y_{|Y|}$ consisting of $|Y|$ characters, the objective is to predict the target token $Y_{kl}=y_k,...,y_l$ that spans $(k,l)$ in the target sentence.

Following the settings in \citet{devlin2019bert} for the SQuAD v2.0 task, the target span can be extracted by predicting the start and end position in the target sentence. The probabilities of the start and end positions of the answer span can be defined as $p_{start}$ and $p_{end}$, respectively. Given the score $w^{X \rightarrow Y}_{ijkl}$ as the product of $p_{start}$ and $p_{end}$, the training objective is to select the answer span $(\hat{k}, \hat{l})$ in the target sentence $Y$ that maximizes the score $w^{X \rightarrow Y}_{ijkl}$, based on the source sentence $X$ and source span $(i, j)$, as shown in the following equations,
\begin{equation}
\label{eq:w}
    w^{X \rightarrow Y}_{ijkl} = p_{start}(k|X,Y,i,j) \times p_{end}(l|X,Y,i,j),
\end{equation}

\begin{equation}
\label{eq:kl}
    (\hat{k}, \hat{l}) = \mathop{\arg \max}\limits_{(k,l):1<k<l<|Y|} {w^{X \rightarrow Y}_{ijkl}}.
\end{equation}

While span prediction works well on word alignment, it still requires datasets with manually aligned parallel sentences. In the following subsections, we propose to pre-train a word alignment model with a large-scale weakly-supervised dataset.
\begin{algorithm}
\caption{Paragraph Pair Collection from Wikipedia}
\label{wikiprocess}
\KwIn{Multilingual paragraph set $\mathcal{P}$ \\
\qquad\quad Language-agnostic entity set $\mathcal{E}$
}

Initialize an empty paragraph pair list $\mathcal{C}$ and
inverted index dictionary $\mathcal{I}$; \\

\ForEach{paragraph $p \in \mathcal{P}$}{
    \tcp*[h]{Get every entity in $p$ by the hyperlink}\\
    $\mathcal{E}_p := GetEntities(p)$;\\
    \ForEach{$e \in \mathcal{E}_p$}
    {
        Append $p$ into $\mathcal{I}[e]$;\\
    }
}

\ForEach{ entity $e \in \mathcal{E}$}{
    Find all paragraphs that mentioned $e$, $\mathcal{P}_e  \subseteq \mathcal{P} := \mathcal{I}[e]$; \\
    Make pair-wise combination for $\mathcal{P}_e$ and append to $\mathcal{C}$; \\
    % \ForEach{ $X \in \mathcal{P}_e$}{
    % \ForEach{ $Y \in \mathcal{P}_e$}{
    % \If{$X$ == $Y$}{Continue;}
    %         \tcp*[h]{Get the positional span that $e$ mentioned with the hyperlink in the paragraph}\\
    %         $(i,j) := GetSpan(X, e)$; \\
    %         $(k,l) := GetSpan(Y, e)$; \\
    %         Add $(X, (i,j), Y, (k,l))$ into $\mathcal{C}$; \\
    % }
    % }
}
\KwOut{Paragraph Pairs with Co-mentioned Entity $\mathcal{C}$}
\end{algorithm}

% \vspace{-1mm}
\subsection{Data Collection and Annotation}
\label{sec:getdata}
% To mitigate the existing approaches' dependency on manual data, we propose to collect weakly-supervised signals from large-scale data on the web. We use Wikipedia data in this paper.
% We broaden the supervision by relaxing the requirements for correct, fully-aligned, parallel sentences. Instead, we collect large-scale noisy (automatically made), partially aligned, non-parallel sentences (or monolingual sentence pairs) for word alignment pre-training.
% \subsubsection{Paragraph Pairs Collection}
Figure~\ref{fig:s3a} shows the framework of our proposed approach.
Firstly, we collect Wikipedia paragraph pairs by co-mentioned hyperlinks.
A typical Wikipedia page contains paragraphs mentioning entities with hyperlinks.
% \nagata{hyperlinks to other pages?}
A hyperlink points to a language-agnostic entity with a unique entity identifier provided by a public project called Wikidata\footnote{\href{https://en.wikipedia.org/wiki/Wikipedia:Wikidata}{https://en.wikipedia.org/wiki/Wikipedia:Wikidata}}. We use those identifiers to build an inverted index dictionary, in which each key is an entity identifier and its corresponding value is a list of paragraphs that mention the entity.
% \nagata{Is it true? I think not all headings has a unique ID in Wikidata because Wikidata does not cover all Wikipedia.}
On the basis of this dictionary, we make two paragraphs as a pair if they are indexed by the same entity, i.e., they contain hyperlinks with the same unique identifier. These two paragraphs can be in any language and on any page. Algorithm~\ref{wikiprocess} elaborates on the collection process.
% \nagata{You have to state Wikidata, which is publicly available, and inverted index, which was made by you.}
% \qiyu{Do we need an example here?}

After obtaining the paragraph pairs, we automatically annotate the word alignments. We categorize words into \textbf{\textit{common words}} and long-tailed \textbf{\textit{wiki words}} and address them separately.

% \vspace{-2mm}
\paragraph{Annotation for Common words}
% In the previous stage, we only make one wiki word alignment for each paragraph pair.
Common words can be defined by existing common word lists\footnote{For example, \href{https://www.wordfrequency.info/}{https://www.wordfrequency.info/}}
% \nagata{Add "For example" before this URL.}
or part-of-speech (POS) tags. In this paper, we use a POS tagger to identify whether a word has a POS tag for common words or not. The common POS tags we used are shown in \S\ref{ap:pos}.
% \nagata{There is no established notion as common POS tags. You have to define what is common POS tags.}
% In this paper, we use a part-of-speech (POS) tagger\footnote{https://github.com/flairNLP/flair} to predict the POS tag for each token in the paragraph and identify tokens with common POS tags as common words.
We take the method in~\citet{JaliliSabet2020SimAlignHQ}, which shows reliable unsupervised ability for word alignment with contextual embeddings in a PLM, to annotate alignments for common words. We make alignments by bi-directional agreement, i.e., two tokens are identified as aligned if they are the most similar token in each other's paragraph. Lastly, we only keep alignments in which at least one of the aligned tokens is common words; otherwise we discard it.

% \vspace{-2mm}
\paragraph{Annotation for wiki words}
A wiki word here denotes a token span in a paragraph. The token span is associated with a hyperlink pointing to an entity, as we introduced in the data collection.
% \nagata{wiki word is not a hyperlink. It is a word that have hyperlink to a Wikidata item.}
% The hyperlinks in the paragraphs pair point to an identical entity as we make the pair in that way.
% Then, we directly use such a co-mention connection to annotate alignments.
% As mentioned above, we collect the paragraph pair, in which the two paragraphs contain an identical entity.
Hence, regardless of what languages in which the wiki words are mentioned, we can make alignments for wiki words by directly aligning the corresponding hyperlinks spans of that co-mentioned entity. 

It is necessary to have separate processes for common words and wiki words because wiki words are mainly named entities, we need alignments for common words to complement them.
It is known that embedding-based methods work well on annotating common word alignments but perform badly for long-tail wiki words as the embeddings of those long-tail words are usually poorly optimized and noisy ~\cite{bahdanau2017learning, gong2018frage,khassanov2019enriching,schick2020rare, Wu2021TakingNO} in a PLM. 

The wiki word and common word alignments are denoted as $\mathcal{D}_{wiki}$ and $\mathcal{D}_{com}$, respectively. The formal definition is given in \S\ref{ap:formal}. After data collection and annotation for wiki words and common words, we combine the two weakly-supervised datasets to obtain the final pre-training dataset, denoted by $\mathcal{D} = \mathcal{D}_{com} \cup \mathcal{D}_{wiki}$.
% \vspace{-1mm}

\subsection{Word Alignment Pre-training via Weakly-Supervised Span Prediction}
\label{sec:train}
\paragraph{Training Objective}
We utilize span prediction as our pre-training objective, as shown in Figure~\ref{fig:s3a}.
As introduced in \S\ref{sec:spanalign}, given a alignment example $(X,Y,i,j,k,l) \in \mathcal{D}$, the objective is to optimize a backbone neural network $f_{\theta^b}$, a start position predictor $g_{\theta^s}$ and an end position predictor $g_{\theta^e}$, which are parameterized by $\theta^b$, $\theta^s$ and $\theta^e$, respectively. The predicted probabilities that $(k,l)$ are the start and end positions of the aligned span in $Y$ can be respectively computed as follows,
\begin{gather}
\begin{aligned}
    &prob(t,\theta^1,\theta^2) = \frac{
        e^{g_{\theta^2}(f_{\theta^1}(X,Y,i,j))_{t}}
    }{
    \sum_{m=1}^{|Y|} e^{g_{\theta^2}(f_{\theta^1}(X,Y,i,j))_m}
    } \\
    &p_{start}(k|X,Y,i,j) = prob(k,\theta^b,\theta^s)
    \\
    &p_{end}(l|X,Y,i,j) = prob(l,\theta^b,\theta^e)
    % &p_{s}(k|X,Y,i,j) =
    % \frac{
    %     e^{g_{\theta^2}(f_{\theta^1}(X,Y,i,j))_k}
    % }{
    % \sum_{m=1}^{|Y|} e^{g_{\theta^2}(f_{\theta^1}(X,Y,i,j))_m}
    % }
    % \\
    % &p_{e}(l|X,Y,i,j) =
    % \frac{
    %     e^{g_{\theta^3}(f_{\theta^1}(X,Y,i,j))_l}
    % }{
    % \sum_{m=1}^{|Y|} e^{g_{\theta^3}(f_{\theta^1}(X,Y,i,j))_m}
    % }
\end{aligned} 
\end{gather}

Then the networks can be applied to $X$, $Y$ and $(i,j)$ to compute the score $w^{X \rightarrow Y}_{ijkl}$ based on Equation~\ref{eq:w}. Following the setting in BERT~\cite{devlin2019bert}, we optimize $\Theta=\{\theta^b,\theta^s,\theta^e\}$ with the following loss for each training example,
\begin{equation}
    L(X,Y,i,i,k,l;\Theta)=
    -\log w^{X \rightarrow Y}_{ijkl}
\end{equation}

% \paragraph{Inference}
% After the pre-training is finished, the model can be directly used to predict word alignments because the training process follows the span prediction way. Given a source sentence $X$, source span $(i,j)$ and target sentence $Y$, the target span $(\hat{k},\hat{l})$ can be predicted by Equations~\ref{eq:w} and \ref{eq:kl}. 
% Hence, \methodname~can also perform \emph{zero-shot} word alignment if we only pre-train the model on monolingual English data. 
% \vspace{-1mm}
\paragraph{Inference and Fine-tuning}
After the pre-training is finished, the model can be directly used to predict word alignments as follows. Given a source sentence $X$, source span $(i,j)$ and target sentence $Y$, the target span $(\hat{k},\hat{l})$ can be predicted by Equations~\ref{eq:w} and \ref{eq:kl}. This setting is denoted as \emph{zero-shot}.
Moreover, our pre-trained model can be easily improved further by fine-tuning on available manual word alignment datasets.
Supervised word alignment is viable because a small amount of gold alignment data can be annotated in hours~\cite{stengel2019discriminative, nagata2020supervised}, which is a reasonable budget in practice if we want to make it perform better on a specific low-resource language pair. 
% We use word alignment datasets introduced in \S\ref{ap:details} for experimental analysis in this paper.
The settings in which a small number and all training examples are used are denoted as \emph{few-shot} and \emph{supervised} fine-tuning, respectively. The experimental settings of few-shot and supervised fine-tuning are the same, except for an increased number of training epochs performed in the few-shot setting. Details are shown in \S\ref{ap:setups}.

\paragraph{Mapping Character-based Prediction to Word Tokens}
As our approach is span-prediction based, the predicted spans may not always align with the original word boundaries. Therefore, following implementation in previous work~\cite{nagata2020supervised}, we select the longest sequence of target tokens that are strictly included in word boundaries in the target sequence as the predicted span. For example, if the model predicts [Yo, \#\#shi, \#\#mits, \#\#u, AS, \#\#HI], we select [Yo, \#\#shi, \#\#mits, \#\#u] as the predicted span because [AS, \#\#HI] is not strictly included in a word.

% \vspace{-1mm}
\paragraph{Symmetric Word Alignment}
The model performs a one-way prediction of the aligned span for the given source tokens. Such an asymmetric prediction can result in inconsistent alignments when we swap the source and target. We follow the strategy in SpanAlign~\cite{nagata2020supervised} to solve it and obtain the final alignment.
Specifically, we can first obtain the token-level alignment probabilities predicted by the model separately in two directions for a pair of sentences. Then, we calculate the symmetric probabilities for each token pair by simply averaging the two probability scores. Lastly, we identify two tokens as aligned if the symmetric probability is larger than a preset threshold.
% In this paper, the threshold is set to 0.4.
\section{Experiments}
% In this section, we will elaborate on the implementation details of our data collection, annotation, and model implementation, followed by various experimental analyses.\nagata{I generally think this kind of outline description can be removed.}
% \subsection{Pre-training Dataset and Benchmark}
\subsection{Pre-training Dataset Details}
\label{sec:dataset}
We pre-train our model in a weakly-supervised manner, in which all pre-training data are automatically collected and annotated in the way described in \S\ref{sec:getdata}.
% In this section, we elaborate on the data and tools used in this paper.\nagata{This can be also deleted.}
We first collect paragraphs from Wikipedia dumps\footnote{\href{https://dumps.wikimedia.org/}{https://dumps.wikimedia.org/}} in English, German, French, Romanian, Chinese and Japanese. Statistics of paragraphs and entities are shown in Table~\ref{tab:rawdata} in Appendix \S\ref{sec:appendix}.
% \nagata{you should clearly indicate that it is in Appendix. Or you might be regarded as exceeding the page limit.}
The connections of inter-language hyperlinks are extracted from Wikidata\footnote{\href{https://en.wikipedia.org/wiki/Help:Interlanguage_links}{https://en.wikipedia.org/wiki/Help:Interlanguage\_links}}. We use \texttt{Wikipedia2Vec}~\footnote{\href{https://wikipedia2vec.github.io/wikipedia2vec/}{https://wikipedia2vec.github.io/wikipedia2vec/}}\cite{yamada2020wikipedia2vec} to extract the paragraphs and co-mention relations of entities.
In this paper, we make the paragraph pairs English-centric, i.e., De-En (German-English), Fr-En (English-Frence), Ro-En (Romanian-English), Zh-En (Chinese-English) and Ja-En (Japanese-English), for more efficient pre-training because most available benchmarks are English-centric. The numbers of sampled examples in each language pair are equal.
% Nevertheless, our proposed data collection and annotation method is applicable to any language pair.

Additionally, we also collect a monolingual dataset in English for testing \methodname's cross-lingual ability, the experimental analysis of which is shown in \S\ref{sec:mono}. The collection process of monolingual data is the same as that of multi-lingual data, except for an additional filter for cross-lingual mentioned entities. That is, we keep only the entities that have been mentioned in another language at least once. We did this for two reasons: one is the explosive computational cost for co-mentions within a language, and we also want entities that appear across various languages because we are testing the cross-lingual alignment ability.

Prior to annotating the alignment, we filter those paragraph pairs by length for more stable training. Specifically, we keep only the pairs with medium length
% \nagata{Does this mean you used sentencepiece? Please describe the number of merge operations.}
, i.e., the pairs that include paragraphs longer than 158 subwords and shorter than 30 subwords are removed. We use SentencePiece with checkpoint \texttt{flores101\_mm100\_615M}~\footnote{\href{https://github.com/flairNLP/flair}{https://github.com/flairNLP/flair} }\cite{goyal2022flores} to tokenize paragraphs in multiple languages, assuming that each sub-word contains a similar amount of information.
After that, we further filter the pairs by semantic similarity because a pair with two unrelated paragraphs is likely to result in no aligned common words between them.
Hence, we keep only the paragraph pairs with a semantic similarity score higher than 0.75, in which the score is calculated by the cosine distance on the embeddings encoded by recent sentence embedding methods. We use \texttt{LaBSE}~\footnote{\href{https://huggingface.co/sentence-transformers/LaBSE}{https://huggingface.co/sentence-transformers/LaBSE}}\cite{feng2022language} and \texttt{pcl-bert-base-uncased}~\footnote{\href{https://github.com/qiyuw/PeerCL}{https://github.com/qiyuw/PeerCL}}\cite{wu2022pcl} as the sentence embedders for multi-lingual and monolingual datasets, respectively.

Lastly, we randomly sample 2,000,000 pairs as the final dataset. As introduced in \S\ref{sec:getdata}, we annotate wiki word alignments for all the 2,000,000 pairs but annotate only randomly selected 200,000 of them for common word alignments. This is because, on average, a paragraph contains more weakly-supervised alignments for common words than wiki words. We use the POS tagger \texttt{flair/upos-multi}~\footnote{\href{https://huggingface.co/flair/upos-multi}{https://huggingface.co/flair/upos-multi}}\cite{akbik2019flair} to identify common words.
% to choose common words from the texts, in which words predicted with common POS tags are regarded as common words.
The statistics in different stages of data collection and annotation are shown in \S~\ref{ap:stat}.
% \begin{figure}
%     \centering
%     \input{Figures/stat.tex}
%     \caption{Statistics of paragraph pairs and alignments
% in the data collection and annotation. ∗We only use
% 200,000 pairs for common word alignment.}
%     \label{fig:stat}
% \end{figure}

% \vspace{-1.5mm}
\subsection{Benchmark Datasets}
% \label{sec:bench}
We evaluate \methodname's performance on five gold word alignment datasets:
% Zh-En, Ja-En, De-En, Ro-En and En-Fr.
Chinese-English (Zh-En), Japanese-English (Ja-En), German-English (De-En), Romanian-English (Ro-En) and English-French (En-Fr).

The Zh-En data is obtained from the GALE Chinese-English Parallel Aligned Treebank~\cite{li2015gale}. We follow \citet{nagata2020supervised} to pre-process the data, in which we use Chinese character-tokenized bitexts, remove mismatched bitexts and time stamps, etc. Then we randomly split the dataset into 80\% for fine-tuning, 10\% for testing and 10\% for future reserves.

The Ja-En data is obtained from the Kyoto Free Translation Task (KFTT)\footnote{\href{http://www.phontron.com/kftt}{http://www.phontron.com/kftt}} word alignment data~\cite{neubig11kftt}. KFTT word alignment data is made by aligning part of the dev and test translation data. We use all eight dev files for fine-tuning, four out of seven test files for testing and the remaining three for future reserves.

The De-En data is from~\citet{vilar2006aer}\footnote{\href{https://www-i6.informatik.rwth-aachen.de/goldAlignment/}{https://www-i6.informatik.rwth-aachen.de/goldAlignment/}}. The Ro-En data and En-Fr data are from the shared task of the HLT-NAACL-2003 Workshop on Building and Using Parallel Texts~\cite{mihalcea2003evaluation}, and the En-Fr data is originally from~\citet{och2003systematic}. We use the pre-processing and scoring scripts\footnote{\href{https://github.com/lilt/alignment-scripts}{https://github.com/lilt/alignment-scripts}} provided by~\citet{zenkel2019adding} for the De-En, Ro-En and En-Fr data, and the number of sentences are 508, 248 and 447, respectively. For De-En and En-Fr, We use 300 sentences for fine-tuning and the remaining for testing. For Ro-En, we use 150 sentences for fine-tuning and the remaining for testing.

\subsection{Experimental Details}
\label{ap:setups}

\paragraph{Pre-training Setups}
We conduct continual pre-training for 100,000 steps with 2,000 warmup steps, starting from multilingual PLMs. We use \texttt{bert-base-multilingual-cased}~\footnote{\href{https://huggingface.co/bert-base-multilingual-cased}{https://huggingface.co/bert-base-multilingual-cased}}~\cite{devlin2019bert} for Zh-En and Ja-En, and \texttt{xlm-roberta-base}~\footnote{\href{https://huggingface.co/xlm-roberta-base}{https://huggingface.co/xlm-roberta-base}} \cite{conneau2020unsupervised} for De-En, En-Fr and Ro-En, respectively. Detailed discussion regarding the choice of PLMs is in \S\ref{sec:bert_xlm}.
We carry out preliminary grid searches on the manual KFTT (Ja-En) training set to decide the hyperparameters. The learning rate is set to 1e-6, the maximum sequence length is set to 384, and the batch size is 96. We use a 12-layer Transformer as the encoder, in which the hidden size is 768, and the number of attention heads is 12.

\paragraph{Fine-tuning Setups}
\label{sec:ftsetup}
For testing the performance on downstream datasets, we fine-tuned the pre-trained model for five epochs for \emph{supervised} and 250 epochs for \emph{few-shot} setting, respectively. The labeled examples we use for \emph{few-shot} is 32.
Following the common practices of pre-training methods, the hyperparameters of fine-tuning are decided empirically by grid-search on the development set. Learning rate is selected from \{1e-6, 3e-6, 1e-5, 3e-5\} and batch size is selected from \{5, 8, 12\}.
Besides, the threshold for symmetric word alignment described in \S\ref{sec:train} is set to 0.4, following SpanAlign~\cite{nagata2020supervised}.

% More detailed information on benchmarks, pre-training and fine-tuning setups are shown in \S\ref{ap:details}.

% \subsection{Competitive Baselines}
% We compare the proposed \methodname~with previous competitive methods including Giza++, SimAlign, AWESomE and SpanAlign. Details are in \S\ref{ap:baseline} For all baselines, we report the best numbers in their original paper.

\subsection{Measures for Word Alignment Quality}
We measure word alignment quality by precision, recall and F1 score in the same way as previous literature~\cite{nagata2020supervised}. Given the predicted alignment results (A), \textit{sure} alignments (S) and \textit{possible} alignments (P). Precision, Recall, and F1 can be calculated as:
\begin{gather}
\begin{aligned}
    &Precision(A,P) = \frac{|A \cap P|}{|A|}\\
    &Recall(A,S) = \frac{|A \cap S|}{|S|}\\
    &F_1 = \frac{2 \times Precision \times Recall}{Precision + Recall}
\end{aligned}
\end{gather}
We also report Alignment Error Rate (AER)~\cite{och2003systematic}, which can be calculated as equation~\ref{eq:aer}, but regard it as a secondary metric because we take the previous literature's~\cite{fraser2007measuring, nagata2020supervised} claim that AER inappropriately favors precision over recall and should be used sparingly. 
\begin{equation}
\label{eq:aer}
    AER(A,S,P) = 1 - \frac{|A \cap S| + |A \cap P|}{|A| + |S|}
\end{equation}
% \nagata{innapropriately favors preision over recall}
Note that only partial word alignment datasets (in our paper, De-En and En-Fr) may distinguish between sure and possible alignments. In the case where \textit{possible} and \textit{sure} alignments are not distinguished (i.e., P == S), AER = 1 - F1. We report both because previous work calculates and reports results in different ways.
In particular, as the En-Fr dataset is known as noisy, special handling was necessary for evaluation in previous studies. And the reported F1 numbers in previous baselines vary greatly due to the different evaluation methods. Consequently, we choose a common practice that fine-tuning on the \textit{sure} data but evaluating on the \textit{sure+possible} data, and we only report AER for En-Fr for a fairer comparison.

\begin{table*}[ht!]
\centering
\resizebox{.9\textwidth}{!}{
\begin{tabular}{l|l|ccc|c}
\hline
Test Set & Method                                                           & Precision    & Recall    & F1            & AER  \\ \hline
Zh-En    & FastAlign~\cite{stengel2019discriminative} & 80.5 & 50.5 & 62.0          &-      \\ 
         & DiscAlign~\cite{stengel2019discriminative} & 72.9 & 74.0 & 73.4          &-     \\
         & SpanAlign~\cite{nagata2020supervised}      & 84.4 & 89.2 & 86.7          & 13.3 \\
         & \methodname~(ours)     & 90.8 & 92.2 & \textbf{91.5 \small{($\uparrow$ 4.8)}} & \textbf{8.5 \small{($\downarrow$ 4.8)}}  \\
         % & \methodname-mBERT$^*$                  & 90.8 & 92.2 & \textbf{91.5} & \textbf{8.5}  \\
         % & \methodname-XLM-R$^*$                        & 83.6 & 91.4 & 87.3          & 12.7 \\
         \hline
Ja-En    & Giza++~\cite{neubig11kftt}                 & 59.5 & 55.6 & 57.6          & 42.4 \\
         & AWESoME~\cite{dou2021word}                 & - & - & -                   & 37.4 \\
         & SpanAlign~\cite{nagata2020supervised}      & 77.3 & 78.0 & 77.6          & 22.4 \\
         & \methodname~(ours)                      & 81.6 & 85.9 & \textbf{83.7 \small{($\uparrow$ 6.1)}} & \textbf{16.3 \small{($\downarrow$ 6.1)}} \\
         % & \methodname-mBERT$^*$                       & 81.6 & 85.9 & \textbf{83.7} & \textbf{16.3} \\
         % & \methodname-XLM-R$^*$                           & 81.2 & 83.8 & 82.5        & 17.5 \\ 
         \hline
De-En    & SimAlign~\cite{JaliliSabet2020SimAlignHQ}      & - & - & 81.0          & 19.0 \\
         & AWESoME~\cite{dou2021word}                 & - & - & -                   & 15.0 \\
         & SpanAlign~\cite{nagata2020supervised}      & 89.9 & 81.7 & 85.6          & 14.4 \\
         % & \methodname-mBERT$^*$                      & 91.9 & 84.9 & 88.3          & 11.7 \\
         % & \methodname-XLM-R$^*$         & 90.7 & 87.1 & \textbf{88.9} & \textbf{11.1} \\ \hline
         & \methodname~(ours)       & 90.7 & 87.1 & \textbf{88.9 \small{($\uparrow$ 3.3)}} & \textbf{11.1 \small{($\downarrow$ 3.3)}} \\ \hline
Ro-En    & SimAlign~\cite{JaliliSabet2020SimAlignHQ}      & - & - & 71.0          & 29.0 \\
         & AWESoME~\cite{dou2021word}                 & - & - & -                   & 20.8 \\
         & SpanAlign~\cite{nagata2020supervised}      & 90.4 & 85.3 & 86.7          & 12.2 \\
         % & \methodname-mBERT$^*$                       & 89.6 & 89.5 & 89.5          & 10.5 \\
         % & \methodname-XLM-R$^*$                & 92   & 90.9 & \textbf{91.4} & \textbf{8.6}  \\ \hline
         & \methodname~(ours)              & 92.0   & 90.9 & \textbf{91.4 \small{($\uparrow$ 4.7)}} & \textbf{8.6 \small{($\downarrow$ 3.6)}}  \\ \hline
En-Fr    & SimAlign~\cite{JaliliSabet2020SimAlignHQ}      & - & - & 93.0          & 7.0 \\
         & AWESoME~\cite{dou2021word}                 & - & - & -                   & 4.1 \\
         & SpanAlign~\cite{nagata2020supervised}      & 97.7 & 93.9 & -           & 4.0 \\
         % & \methodname-mBERT$^*$                   & 80.5 & 95.5 & 87.4          & 10.5 \\
         % & \methodname-XLM-R$^*$                    & 81.5 & 96.0  & 88.2 & 11.8 \\
         & \methodname~(ours)                  & 98.8 & 96.0  & - & \textbf{2.5 \small{($\downarrow$ 1.5)}} \\
\hline
\end{tabular}
}
\caption{Comparison of \methodname~and previous methods on word alignment datasets. Higher F1 scores are better. Lower AER scores are better. We highlight the best number in the same setting and test set with bold font.}
% \vspace{-3mm}

\label{tab:main}
\end{table*}
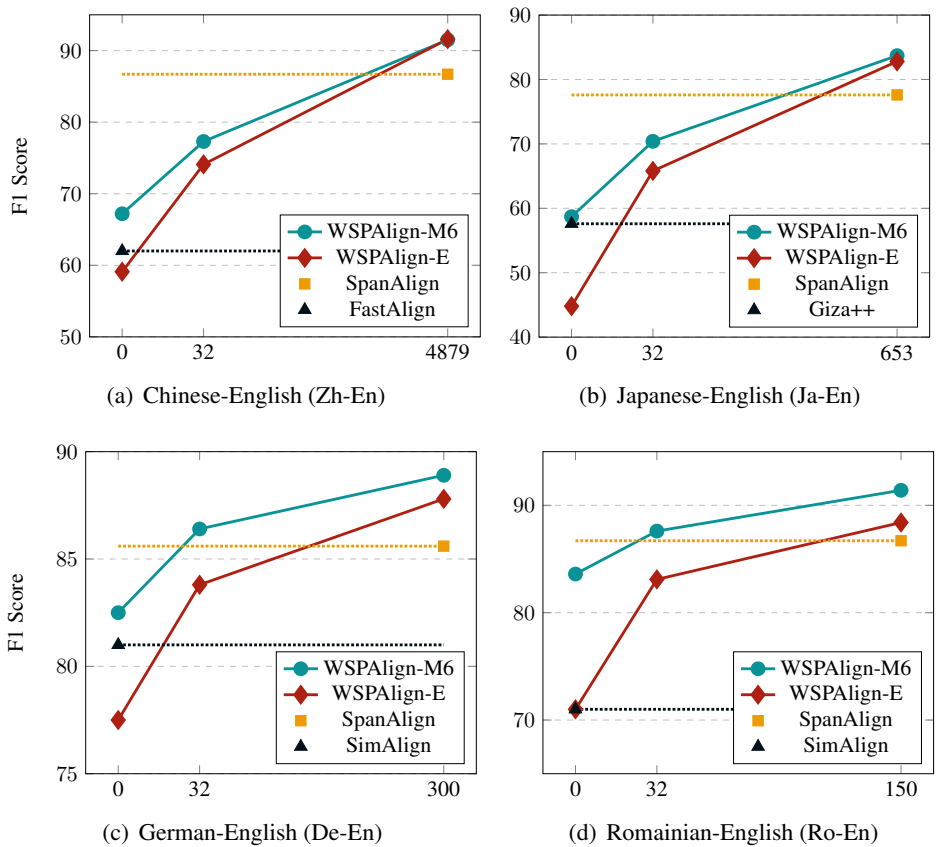
\begin{figure*}[t]
    \centering
% \resizebox{1.5\textwidth}{!}{
    
    \subfigure[Chinese-English (Zh-En)]{
        \begin{tikzpicture}[scale=.75]
\begin{axis}[
    ylabel={F1 Score},
    xtick=data,
    xticklabels={0,32,4879},
    ymin=50,
    ymax=95,
    legend pos=south east,
    ymajorgrids=true,
    grid style=dashed,
]

\addplot[
    color=color1,
    line width=0.5mm,
    mark=*,
    mark options={scale=1.5, fill=color1},
    ]
    coordinates {
    (0,67.2)(1,77.3)(4,91.5)
    };

\addplot[
    color=color2,
    line width=0.5mm,
    mark=diamond*,
    mark options={scale=2, fill=color2},
    ]
    coordinates {
    (0,59.1)(1,74.1)(4,91.6)
    };

\addplot[
    color=color3,
    only marks,
    mark=square*,
    mark options={scale=1.3, fill=color3},
    ]
    coordinates {
    (4,86.7)
    };
\addplot[
    color=color4,
    only marks,
    mark=triangle*,
    mark options={scale=1.8, fill=color4},
    ]
    coordinates {
    (0,62.0)
    };
    
\addplot[
    color=color3,
    line width=0.5mm,
    no marks,
    densely dotted,
    ]
    coordinates {
    (0,86.7)(4,86.7)
    };
\addplot[
    color=color4,
    line width=0.5mm,
    no marks,
    densely dotted,
    ]
    coordinates {
    (0,62.0)(4,62.0)
    };
    
\addlegendentry{\methodnamesix}
\addlegendentry{\methodnameone}
\addlegendentry{SpanAlign}
\addlegendentry{FastAlign}
    
\end{axis}
\end{tikzpicture}
        \label{fig:few1}
    } \hspace{-2mm}
    \subfigure[Japanese-English (Ja-En)]{
        \begin{tikzpicture}[scale=.75]
\begin{axis}[
    xtick=data,
    xticklabels={0,32,653},
    ymin=40,
    ymax=90,
    legend pos=south east,
    ymajorgrids=true,
    grid style=dashed,
]

\addplot[
    color=color1,
    line width=0.5mm,
    mark=*,
    mark options={scale=1.5, fill=color1},
    ]
    coordinates {
    (0,58.7)(1,70.4)(4,83.7)
    };
\addplot[
    color=color2,
    line width=0.5mm,
    mark=diamond*,
    mark options={scale=2, fill=color2},
    ]
    coordinates {
    (0,44.8)(1,65.8)(4,82.8)
    };

\addplot[
    color=color3,
    only marks,
    mark=square*,
    mark options={scale=1.3, fill=color3},
    ]
    coordinates {
    (4,77.6)
    };
\addplot[
    color=color4,
    only marks,
    mark=triangle*,
    mark options={scale=1.8, fill=color4},
    ]
    coordinates {
    (0,57.6)
    };

\addplot[
    color=color3,
    line width=0.5mm,
    no marks,
    densely dotted,
    ]
    coordinates {
    (0,77.6)(4,77.6)
    };
\addplot[
    color=color4,
    line width=0.5mm,
    no marks,
    densely dotted,
    ]
    coordinates {
    (0,57.6)(4,57.6)
    };

\addlegendentry{\methodnamesix}
\addlegendentry{\methodnameone}
\addlegendentry{SpanAlign}
\addlegendentry{Giza++}
    
\end{axis}
\end{tikzpicture}
        \label{fig:few2}
    } \hspace{-2mm}
    \subfigure[German-English (De-En)]{
        \begin{tikzpicture}[scale=.75]
\begin{axis}[
    ylabel={F1 Score},
    xtick=data,
    xticklabels={0,32,300},
    ymin=75,
    ymax=90,
    legend pos=south east,
    ymajorgrids=true,
    grid style=dashed,
]

\addplot[
    color=color1,
    line width=0.5mm,
    mark=*,
    mark options={scale=1.5, fill=color1},
    ]
    coordinates {
    (0,82.5)(1,86.4)(4,88.9)
    };
\addplot[
    color=color2,
    line width=0.5mm,
    mark=diamond*,
    mark options={scale=2, fill=color2},
    ]
    coordinates {
    (0,77.5)(1,83.8)(4,87.8)
    };

\addplot[
    color=color3,
    only marks,
    mark=square*,
    mark options={scale=1.3, fill=color3},
    ]
    coordinates {
    (4,85.6)
    };
\addplot[
    color=color4,
    only marks,
    mark=triangle*,
    mark options={scale=1.8, fill=color4},
    ]
    coordinates {
    (0,81.0)
    };

\addplot[
    color=color3,
    line width=0.5mm,
    no marks,
    densely dotted,
    ]
    coordinates {
    (0,85.6)(4,85.6)
    };
\addplot[
    color=color4,
    line width=0.5mm,
    no marks,
    densely dotted,
    ]
    coordinates {
    (0,81.0)(4,81.0)
    };
    
\addlegendentry{\methodnamesix}
\addlegendentry{\methodnameone}
\addlegendentry{SpanAlign}
\addlegendentry{SimAlign}
    
\end{axis}
\end{tikzpicture}
        \label{fig:few3}
    }
    \subfigure[Romainian-English (Ro-En)]{
        \begin{tikzpicture}[scale=.75]
\begin{axis}[
    xtick=data,
    xticklabels={0,32,150},
    ymin=65,
    ymax=95,
    legend pos=south east,
    ymajorgrids=true,
    grid style=dashed,
]

\addplot[
    color=color1,
    line width=0.5mm,
    mark=*,
    mark options={scale=1.5, fill=color1},
    ]
    coordinates {
    (0,83.6)(1,87.6)(4,91.4)
    };
\addplot[
    color=color2,
    line width=0.5mm,
    mark=diamond*,
    mark options={scale=2, fill=color2},
    ]
    coordinates {
    (0,71.0)(1,83.1)(4,88.4)
    };

\addplot[
    color=color3,
    only marks,
    mark=square*,
    mark options={scale=1.3, fill=color3},
    ]
    coordinates {
    (4,86.7)
    };
\addplot[
    color=color4,
    only marks,
    mark=triangle*,
    mark options={scale=1.8, fill=color4},
    ]
    coordinates {
    (0,71.0)
    };

\addplot[
    color=color3,
    line width=0.5mm,
    no marks,
    densely dotted,
    ]
    coordinates {
    (0,86.7)(4,86.7)
    };
\addplot[
    color=color4,
    line width=0.5mm,
    no marks,
    densely dotted,
    ]
    coordinates {
    (0,71.0)(4,71.0)
    };
    
\addlegendentry{\methodnamesix}
\addlegendentry{\methodnameone}
\addlegendentry{SpanAlign}
\addlegendentry{SimAlign}
    
\end{axis}
\end{tikzpicture}
        \label{fig:few4}
    }
    % }
    % \vspace{-1mm}
    \caption{Comparison of varying scales of manual samples used on four word alignment test sets. The y-axis is F1 score and the x-axis is the number of manual samples used in the training.
    % \qiyu{reimplemnet simalign}
    }
% \vspace{-3mm}
    \label{fig:few}
\end{figure*}
\subsection{Main Quantitative Results}
\label{sec:mainres}
In this section, we use all available training examples in the benchmark datasets to reach the best potential of \methodname~in the \emph{supervised} fine-tuning setting. The competitive baselines include Giza++, SimAlign, AWESoME and SpanAlign, whose details are described in \S\ref{sec:related}. For baselines, we report the best numbers in their original paper.

Table~\ref{tab:main} shows the comparison of our proposed model and existing approaches. It demonstrates that \methodname~significantly outperforms all supervised and unsupervised baselines. Specifically, \methodname~ improves the best supervised baseline by \textbf{3.3\textasciitilde 6.1} points in F1 and \textbf{1.5\textasciitilde 6.1} points in AER.
% \nagata{absolute percentage points are better}
% Specifically, \methodname~improves the supervised baseline with a large margin on the F1 score 5.5\% and 7.8\% on Zh-En and Ja-En, respectively. For De-En and Ro-En, \methodname~also improves the supervised baseline by 3.8\% and 5.4\%. 
% Besides the F1 score, 
% \methodname~also improves the supervised baseline from 13.3 to 8.5 and 22.4 to 16.3 on Zh-En and Ja-En on AER, respectively. For De-En, Ro-En and En-Fr, the improvements are from 14.4 to 11.1, 12.2 to 8.6 and 4.0 to 2.5, respectively.

Additionally, we can observe that \methodname~improves the baselines on Ja-En with a relatively larger margin. As Japanese is known as a language distant from English, this indicates \methodname's superiority in word alignment in difficult language pairs by introducing more cross-lingual information in the pre-training.

\subsection{Zero-shot and Few-shot Performance}
With varying scales of manual training examples used after the pre-training, we evaluate the zero-shot and few-shot performance of \methodname. As shown in Figure~\ref{fig:few},
% The x-axis is the number of examples used for fine-tuning, in which
we test 0 (\textit{zero-shot}), 32 (\textit{few-shot}) and the full amount (\textit{supervised}) of examples in the benchmark datasets. Details regarding the implementation can be found in \S\ref{sec:train} and \S\ref{sec:ftsetup}.

The circle points with the green line show the performance trend of \methodname~pre-trained on weakly supervised data in six languages (\methodnamesix~in Figure~\ref{fig:few}).
% Details of the dataset are in \S\ref{sec:dataset}. 
For all test sets, zero-shot \methodnamesix~outperforms the unsupervised baselines, and the few-shot \methodnamesix~with only 32 training examples significantly outperforms the unsupervised baselines by a large margin.
This indicates that the proposed pre-training method has a basic zero-shot word alignment ability with no need for any manual data, and the performance can be further improved with only a small number of training examples.

Notably, zero-shot \methodnamesix~beats the unsupervised baselines by a large margin and almost reaches the performance of the supervised baseline on Ro-En. On Ro-En and De-En, \methodnamesix~even slightly outperforms the fully supervised baseline.
As English is known to be closer to Romanian and German than Chinese and Japanese, the results imply that the proposed approach has a higher reward when the downstream languages to be aligned are close. Additionally, the Ro-En and De-Rn datasets respectively include only 150 and 300 training examples, which can make the supervised methods not perform satisfactorily. Thus, considering the computation cost of the pre-training in practice, our proposed large-scale span prediction pre-training with weakly supervised data can bring more benefits in the case when available manual data are scarce or the downstream languages are close.

\section{Discussion}
\subsection{Mono-lingual Span Prediction Pretraining}
\label{sec:mono}
In this section, we will examine mono-lingual span prediction pretraining by pre-training on English-only data but testing on other languages, to investigate the potential cross-lingual ability of \methodname~to confirm whether it is ready for practical application.
Although Wikipedia and recent multi-lingual PLMs support hundreds of languages, the amount of information available for minority languages can still be small~\footnote{\href{https://meta.wikimedia.org/wiki/List\_of\_Wikipedias}{https://meta.wikimedia.org/wiki/List\_of\_Wikipedias}}. How to address such language equality problems is often discussed in recent NLP research~\cite{conneau2020unsupervised,costa2022no}.
In the scope of this paper, even if we collect supervision signals from large-scale encyclopedias and PLMs, the datasets could still be limited for exceptionally low-resource languages in practice.
% \nagata{It is better to move this paragraph to Introduction.}

The diamond points with the red line in Figure~\ref{fig:few} show the performance trend of \methodname~pre-trained on English-only alignment data, i.e., \methodnameone. We observe that zero-shot \methodnameone~underperforms the unsupervised baseline, except on the easier Ro-En test set. However, \methodnameone~can be significantly improved and outperforms the existing unsupervised baselines with only 32 manual examples, which can be collected at a low cost. If we further fine-tune \methodnameone~with a full supervised dataset, it can outperform the supervised baseline on all test sets.
These observations show that with only pre-training on monolingual weakly supervised alignments, \methodname~is not able to be a better word aligner than the existing ones, although it achieves a basic zero-shot ability. However, fine-tuning it on a small number of manual examples can be a practical cross-lingual word aligner better than unsupervised baselines. Moreover, it can beat the state-of-the-art method when the same amount of manual examples are available.

Such a cross-lingual transferring ability that holds for zero-shot, few-shot, and supervised settings suggests that \methodname~is potentially very practical for low-resource languages by only pre-training on large-scale monolingual data, as low-resource language resources are always hard to collect.

From another perspective, our proposed \methodname~consists of two components: span prediction and bilingual equivalence identification. As an ablation study of \methodname, mono-lingual span prediction pre-training performs without bilingual equivalence knowledge but only learns the span prediction. Intriguingly, mono-lingual span prediction still improves bilingual word alignment accuracy in the above experiments. A possible explanation for this result is that word embeddings are somehow aligned out of the box in a multilingual language model. This indicates that only optimizing on mono-lingual span prediction in our proposed method can also potentially generalize to cross-lingual word alignment.

% \paragraph{ft with monolingual manual data}

% \vspace{-2mm}
\subsection{Effect of Common Words and Wiki Words}
\begin{table}[]
\centering
\resizebox{\columnwidth}{!}{
\begin{tabular}{l|ccc|c}
\hline
                           & P & R & F1 & AER \\ \hline
SpanAlign                  &  84.4 &  89.2 & 86.7 &  13.3 \\ \hline
\methodname &  90.8  &  92.2  &  91.5  & 8.5 \\
~~~~w/o common words  &  91.3  &  85.4   &  88.3  &   11.7  \\
~~~~w/o Wiki words  &    91.5   &  86.0   &  88.6  &  11.4 \\ \hline
\end{tabular}
}
\caption{Ablation study by removing common words or wiki items for alignment. Performance on Zh-En test set. Higher F1 is better and lower AER is better.}
% \vspace{-1mm}
\label{tab:ab}
\end{table}
% As introduced in \S\ref{sec:method} we categorize words in the raw texts into common words and wiki words with different methods to make the weakly supervised alignments. In this section,\nagata{You can remove the first sentence and up to here.}
We test two variants of \methodname~by removing the common words and wiki words in the pre-training data, i.e., \methodname~w/o common and \methodname~w/o Wiki. We chose the largest benchmark dataset Zh-En and the setting of supervised fine-tuning for testing.
The experimental results in Table~\ref{tab:ab} show that when alignments for common words or wiki words are removed from the training data, the performance of \methodname~will drop by about 3 points on F1 and AER. But both two variants outperform the supervised baseline SpanAlign. This indicates that the improvement from our proposed weakly supervised pre-training still holds even when we make alignments only for either common words or wiki words, and using both leads to better performance.

% convergence~\ref{fig:converge}
% \begin{figure}[]
%     \centering
%     \subfigure[onlysim]{
%         \includegraphics[width=.35\textwidth]{Figures/onlysim.png}
%         \label{fig:onlysim}
%     } \hspace{-2mm}
%     \subfigure[onlywiki]{
%         \includegraphics[width=.35\textwidth]{Figures/onlywiki.png}
%         \label{fig:onlywiki}
%     } \hspace{-2mm}
%     \caption{convergence of onlywiki and onlysim \qiyu{the figure is just temporal placeholder}}
%     \label{fig:converge}
% \end{figure}

\subsection{The Choice of Multi-lingual PLMs}
\label{sec:bert_xlm}
\begin{table}[]
\centering
\resizebox{\columnwidth}{!}{
\begin{tabular}{c||rrr|r||rrr|r}
\hline
\multirow{2}{*}{Test Set} & \multicolumn{4}{c||}{mBERT} & \multicolumn{4}{c}{XLM-R} \\ \cline{2-9} 
 & P & R & F1 & AER & P & R & F1 & AER \\ \hline
Zh-En & 90.8 & 92.2 & \cellcolor{yellow}91.5 & \cellcolor{yellow} 8.5 & 83.6 & 91.4 & 87.3 & 12.7 \\
Ja-En & 81.6 & 85.9 & \cellcolor{yellow}83.7 & \cellcolor{yellow}16.3 & 81.2 & 83.8 & 82.5 & 17.5 \\
De-En & 91.9 & 84.9 & 88.3 & 11.7 & 90.7 & 87.1 & \cellcolor{yellow}88.9 & \cellcolor{yellow}11.1 \\
Ro-En & 89.6 & 89.5 & 89.5 & 10.5 & 92 & 90.9 & \cellcolor{yellow}91.4 & \cellcolor{yellow}8.6 \\
% En-Fr & - & - & - & - & 98.8 & 96.0 & - & \cellcolor{yellow}2.5 \\ 
\hline
\end{tabular}
}
\caption{The performance on the test sets with different PLM initiation. We highlight the better performance in the same setting with the yellow box.}
% \vspace{-2mm}
\label{tab:bertxlm}
\end{table}
Besides the span prediction pre-training we propose, \methodname~still needs a prior conventional language pre-training to ensure the basic ability of language understanding. In this paper, we start the span prediction pre-training from the checkpoint of two popular multi-lingual PLMs, mBERT~\cite{devlin2019bert} and XLM-R~\cite{conneau2020unsupervised}.
To investigate the effect of different PLMs used, we compare the performance of \methodname~with mBERT and XLM-R on all test sets except En-Fr. We do not use En-Fr because the dataset is noisy.
Table~\ref{tab:bertxlm} clearly shows that mBERT performs better on Zh-En and Ja-En. In contrast, XLM-R performs better on De-En and Ro-En.  Such a difference in performance may be caused by the tokenization method during the language pre-training. The byte-level sub-word tokenization used in RoBERTa~\cite{liu2019roberta} can work poorly for Chinese and Japanese because the character is the smallest unit in these languages.
Hence we use mBERT as the initialization checkpoint for Zh-En and Ja-En and XLM-R for the rest. We also suggest choosing the appropriate PLM for \methodname~according to the downstream languages in practice.

% \subsection{use all available manual data to ft the multi/mono \methodname}
% \vspace{-2mm}
\section{Conclusion}
% \vspace{-2mm}
In this paper, we propose to pre-train word aligners with weakly-supervised signals that can be automatically collected.
% The proposed span prediction pre-pretraining does not require any parallel sentences and annotated alignment examples. Thus it can well mitigate the dependence on manual data when learning word aligners.
We broaden the source of supervision by relaxing the requirement for correct, fully-aligned, and parallel sentences. Specifically, we make noisy, partially aligned, and non-parallel paragraphs on a large scale. 
% We conduct extensive experiments on standard benchmarks in six languages. When fine-tuning it on training examples, the proposed method outperforms the supervised baseline. The few-shot and zero-shot variants also show competitive performance.
% Additionally, we also test the cross-lingual ability of the proposed method by pre-training it on monolingual data.
% the zero-shot model underperforms the baseline, yet the few-shot model with only 32 examples outperforms the unsupervised baseline.
Experimental results in this paper show that pre-training with large-scale weakly-supervision can significantly improve existing word alignment methods and make word aligners more practical as well because no manual data is needed.
% Since annotating alignment data at scale is usually costly,
We provide empirical evidence of how much large-scale span prediction pre-training can help word alignment in terms of data accessibility, the number of manual examples used, and cross-lingual ability.
We hope this paper can contribute to further exploiting practical word alignment techniques with large-scale weak supervision.

\section*{Limitations}
Although \methodname~successfully outperforms all existing baselines, it is still limited to the accessibility of low-resource language information. For example, the collection of pre-training data requires multi-lingual POS tagging tools to identify which words are common or not. It also requires a multi-lingual PLM and Wikipedia hyperlinks to make the alignments, which could be inaccessible for an exceptional minority language. But note that we showed \methodname's cross-lingual ability in \S\ref{sec:mono}, which implies that this issue can potentially be addressed in the direction of pre-training on large-scale monolingual data with our future effort. Besides, this paper lacks evaluation on real low-resource language benchmarks because there is no existing test set. We will try to collect and annotate low-resource word alignment data in our future work.

\section*{Ethics Statement}
This paper investigates the pre-training for word alignment, which will not lead to a negative social impact. The data used in this paper are all publicly available and are widely adopted in previous literature, avoiding any copyright concerns. The proposed method does not introduce ethical bias. On the contrary, our aim is to advance word alignment techniques to enhance their utility for low-resource language communities, promoting inclusivity and equitable access to language resources.

\section*{Acknowledgement}
We thank Ryokan Ri for the valuable discussion and assistance with Wikipedia2vec. Qiyu Wu was supported by JST SPRING, Grant Number JPMJSP2108.

% \clearpage

% \section*{Acknowledgements}
% Thank ryokan

\bibliography{anthology,custom}
\bibliographystyle{acl_natbib}

\clearpage
\appendix
\begin{table}[!bp]
\centering
\begin{tabular}{l|l}
\hline
POS Tag & Meaning                                                  \\ \hline\hline
ADJ     & adjective                                                \\ \hline
VERB    & verb                                                     \\ \hline
DET     & determiner                                               \\ \hline
ADP     & adposition                                               \\ \hline
AUX     & auxiliary \\ \hline
PRON    & pronoun                                                  \\ \hline
PART    & particle                                                 \\ \hline
SCONJ   & subordinating conjunction                                \\ \hline
NUM     & numeral                                                  \\ \hline
NOUN    & noun      \\ \hline
ADV     & adverb                                                   \\ \hline
CCONJ   & coordinating conjunction                                 \\ \hline
INTJ    & interjection                                             \\ \hline
\end{tabular}
\caption{The Meaning of POS tags.}
\label{tab:pos}
\end{table}
\begin{table}[!ht]
\centering
\begin{tabular}{l|r|r}
\hline
   & \# of entities & \# of paragraphs \\ \hline
Zh & 1,768,012      &     22,409,574     \\ \hline
En & 8,675,433      &    145,441,685     \\ \hline
Ja & 1,663,517      &     51,377,620     \\ \hline
Ro & 754,005        &     7,105,064     \\ \hline
De & 3418485        &    57,121,818   \\ \hline
Fr & 3507481        &    63,551,555     \\ \hline
\end{tabular}
\caption{Statistics of Wikipedia raw data.}
\label{tab:rawdata}
\end{table}
\begin{table}[!ht]
\resizebox{\columnwidth}{!}{
\centering
\begin{tabular}{l|r|r}
\hline\hline
\multicolumn{3}{c}{\# of paragraph pairs} \\ \hline
 & Multi-lingual & Monolingual \\ \hline
with co-mention     & 89,973,019    &  72,677,385  \\  
~~-- filter by length        & 41,418,902    & 40,759,166    \\ 
~~-- filter by similarity    & 10,016,210    &  11,304,002   \\  
~~-- finally used            & 2,000,000     & 2,000,000   \\ \hline\hline
\multicolumn{3}{c}{\# of alignment annotations} \\ \hline
 & Multi-lingual & Monolingual \\ \hline
wiki items  & 2,000,000  &    2,000,000   \\ 
common words$^*$ & 1,644,019   & 2,591,357   \\ \hline
\end{tabular}
}
\caption{Statistics of paragraph pairs and alignments in the data collection and annotation. $^*$We use only 200,000 pairs for common word alignment.}
% \vspace{-4mm}
\label{tab:datastat}
\end{table}
\begin{table*}[!t]
\centering
\resizebox{\textwidth}{!}{
\begin{tabular}{c|c|c|c|c}
\hline
Setting                                 & GPU                                      & Dataset      & \# of Training Examples & Training Time (hours)\\ \hline\hline
\multirow{2}{*}{Pre-training}           & \multirow{2}{*}{NVIDIA Tesla A100 (80G)} & 6 languages  & 2,000,000               & 40            \\ \cline{3-5} 
                                        &                                          & English only & 2,000,000               & 42            \\ \hline
\multirow{5}{*}{Supervised Fine-tuning} & \multirow{5}{*}{NVIDIA Titan Xp (12G)}    & Zh-En        & 4,879                   & 6             \\ \cline{3-5} 
                                        &                                          & Ja-En        & 653                     & 3             \\ \cline{3-5} 
                                        &                                          & De-En        & 300                     & 1             \\ \cline{3-5} 
                                        &                                          & Ro-En        & 150                     & 0.25          \\ \cline{3-5} 
                                        &                                          & En-Fr        & 300                     & 1             \\ \hline
Few-Shot Fine-tuning                    & NVIDIA Titan Xp (12G)                     & -$^*$           & 32                      & 2             \\ \hline
\end{tabular}
}
\caption{Experimental environments and training time.$^*$ Training time for each dataset in the few-shot setting is approximately equal.}
\label{tab:traintime}
\end{table*}

\section{Appendix}
\label{sec:appendix}
\subsection{Formal Definition of Annotation for Alignments}
\label{ap:formal}
\paragraph{Wiki Words}
Given a paragraph pair $X$ and $Y$, $X$ and $Y$ contain an identical entity $e$. Suppose $(i,j)$ and $(k,l)$ are the spans
\footnote{The explicit text of the spans can be different, but they refer to the same entity.}
of $e$ in $X$ and $Y$ respectively, we add the alignment of $(X_{ij}, Y_{kl})$ into the dataset $\mathcal{D}_{wiki}$.
\paragraph{Common Words}
Assume we have a parameterized network $\delta$ (e.g., a PLM) that can be applied to a token $X_{ij}$ in the paragraph to derive a dense real-valued vector $\mathbf{h}^X_{ij}=\delta(X_{ij}) \in \mathbb{R}^d$. Then we can calculate the similarity scores for the embedding of each token in the source paragraph $X$ and target paragraph $Y$, and obtain pairwise similarity scores $S$ for every token in the paragraph pairs,
$
    S_{ijkl}^{X \rightarrow Y} = \mathop{sim}(\mathbf{h}^X_{ij}, \mathbf{h}^Y_{kl}),    
$
where $\mathop{sim}$ is a similarity function for two vectors, e.g., cosine similarity.
Then, for two words $(i,j)$ in source sentence $X$ and $(k,l)$ in target sentence $Y$, we annotate the alignment of $(X_{ij}, Y_{kl})$ if and only if
$
((i,j)= \mathop{\arg \max}\limits_{(i,j):1<i<j<|X|}{S_{ijkl}^{X \rightarrow Y}}) \wedge ((k,l)= \mathop{\arg \max}\limits_{(k,l):1<k<l<|Y|}{S_{klij}^{Y \rightarrow X}}).
$
% \begin{equation}
% A_{ij}^{kl}=\left\{\begin{matrix} 1,((i,j)= \mathop{\arg \max}\limits_{(i,j) \in X}{S_{ij}^{kl}}) \wedge ((k,l)= \mathop{\arg \max}\limits_{(k,l) \in Y}{S_{ij}^{kl}}) \\ 
% 0, otherwise
% \end{matrix}\right.
% \end{equation}
As we mentioned earlier, embedding-based methods can perform badly on rare words. Thus we further filter out alignments with common words. That is, given an annotated alignment $(X_{ij}, Y_{kl})$, we add it into the dataset $\mathcal{D}_{com}$ if $(i,j)$ or $(k,l)$ is a common word. Otherwise, we discard it.

\subsection{Common POS Tags}
\label{ap:pos}
We use tags shown in Table~\ref{tab:pos} as common tags~\footnote{\href{https://huggingface.co/flair/upos-multi}{https://huggingface.co/flair/upos-multi}}. Tokens predicted as one of these tags are identified as common words in our method. 

\subsection{Statistics of datasets}
\label{ap:stat}
Table~\ref{tab:rawdata} shows the statistics of Wikipedia raw data we use. English has the most numbers of paragraphs and entities, while Romanian has the least paragraphs and entities. Besides, we also count the number of paragraph pairs and alignment annotations in different phases while obtaining the pre-training data. Specific statistics is shown in the Table~\ref{tab:datastat}.

\subsection{Experimental Enviroments}Table~\ref{tab:traintime} shows the experimental environments and training hours in different settings. 
We used two NVIDIA Tesla A100 (80G) to conduct the pre-training. The pre-training time is around 40 hours. We used Titan X (12G) to conduct the few-shot and supervised fine-tuning, which can be finished in hours for each run. Note that the few-shot fine-tuning has fewer examples but performs 250 epochs, while supervised fine-tuning only performs for 5 epochs.

% This is an appendix.

\end{document}